\documentclass{article}
\usepackage{amsmath,graphicx}
\usepackage[preprint]{spconf}
\usepackage{times}
\usepackage{epsfig}
\usepackage{graphicx}
\usepackage{amsmath}
\usepackage{amssymb}
\usepackage{subcaption}
\usepackage{multirow}
\usepackage[numbers]{natbib}
\usepackage[]{bm}

\usepackage{pifont}
\newcommand{\cmark}{\ding{51}}%
\newcommand{\xmark}{\ding{55}}%

\title{Normal Similarity Network for Generative Modelling}
\name{Jay Nandy, Wynne Hsu, Mong Li Lee}
\address{School of Computing, National University of Singapore} 

\copyrightnotice{\copyright\ IEEE 2018}
\toappear{To appear in Proc.IEEE ICIP, 2018, Athens, Greece}

\begin{document}
%
\maketitle
\begin{abstract}
Gaussian distributions are commonly used as a key
building block in many generative models. 
However, their applicability has not been well explored in deep networks. 
In this paper, we propose a novel deep generative model named
as Normal Similarity Network (NSN) where the layers are 
constructed with Gaussian-style filters.
NSN is trained with a layer-wise 
non-parametric density estimation 
algorithm that iteratively down-samples 
the training images and captures the density 
of the down-sampled training images in the 
final layer.
Additionally, we propose NSN-Gen for generating
new samples from noise vectors by iteratively reconstructing 
feature maps in the hidden layers of NSN.
Our experiments suggest encouraging results of the proposed 
model for a wide range of computer vision applications 
including image generation, styling 
and reconstruction from occluded images.
\end{abstract}

\section{Introduction}

Unsupervised learning, where no labels are provided during training,
remains one of the core challenges in machine learning.
One branch of unsupervised learning that is popular in  
image processing/computer vision is the generative models
that aim to obtain a representation of the 
density function, $p_{model}$ to describe the density of a given 
data domain, $p_{data}$. 
Auto-regressive models such as Fully Visible Belief Nets \cite{fvbn}, PixelRNN \cite{pixelRNN}, explicitly model the joint distribution of 
pixels as a product of conditional distributions and optimize the likelihood of training data.
However, due to higher dimension and the structured formation of images, modelling the long-range pixel correlation becomes challenging for these explicit tractable density models.
Instead of designing an explicit tractable model, 
Variational Autoencoder (VAE) \cite{vae1,vae2}) approximates the 
likelihood by maximizing a lower bound.
This is achieved with an encoder-decoder neural network architecture. 
The encoder's output fits a prior distribution by minimizing the KL-divergence. 
The decoder then transforms encoder's output to reconstruct the images. 
In practice, they often achieve higher likelihood values, however, fails 
to generate sharp images \cite{deeplearningbook}.
\begin{figure}[htbp]
\center
\includegraphics[scale=0.4]{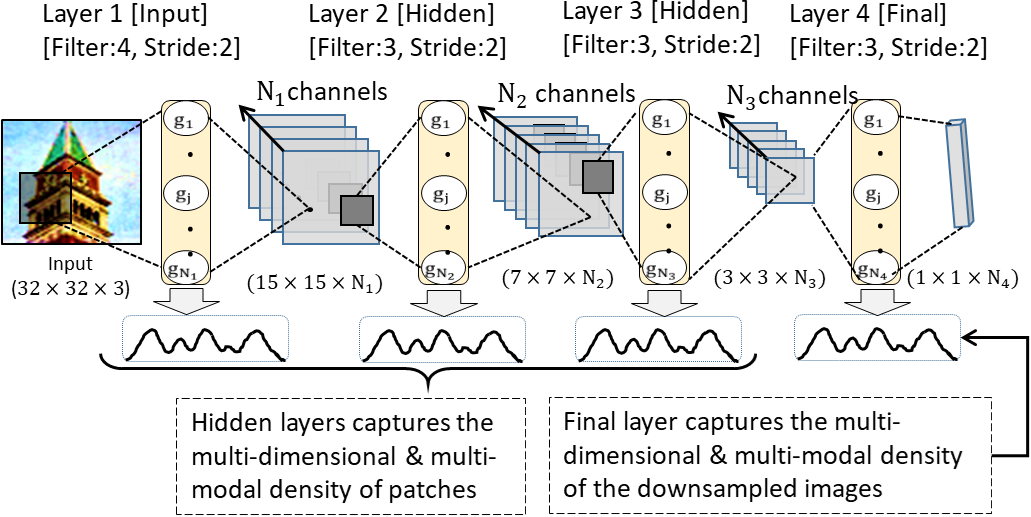}
\caption{Schematic diagram of a 4-layered NSN. 
Each layer of NSN captures the density of 
patches via likelihood maximization criterion and downsamples the images.
After down-sampling the training images, 
the final layer captures the density of them in a transformed domain.
}
\label{fig:NSN}
\end{figure}

Generative Adversarial Networks (GANs) offer an alternative solution that draw samples from $p_{data}$ to learn an implicit density 
function through a minimax formulation \cite{gan2014,nipsGan,dcgan2016,lsgan,wgan}.
However, GANs often suffer from training instability  \cite{ganProblem1}. 
Although recent progresses for GANs aim to address this issue by  
replacing the Jensen-Shannon divergence (JSD) \cite{fgan,wgan,lsgan},
 the new formulations generally can not address the mode collapsing problem where the generator can only accommodate 
a few  modes in the training domain \cite{nipsGan}. 
Tricks such as feature matching, mini-batch discrimination \cite{improvedGan} or by unrolling multiple steps during the 
gradient update in training \cite{unrolledGAN}, handle this issue to some extent.


In this paper, we  develop a deep generative 
model that shares the properties of both explicit density models such as PixelRNN, VAE as well as implicit density models such as GAN. 
We propose a Normal Similarity Network (NSN) with multiple hidden layers where every layer is constructed with learnable Gaussian-style filters.
Figure \ref{fig:NSN} presents a schematic diagram of NSN.
The proposed model downsamples the images and explicitly captures the density in a 
transformed domain via likelihood maximization.
Further, we propose a sampling approach, NSN-Gen, to generate new images from NSN.
NSN-Gen takes a noise vector as input at the final layer and performs a backward pass to generate the pixels in parallel. 
In our experiments, we demonstrate the applicability of NSN 
to  a wide range of tasks including image generation, image styling and reconstruction from occluded images. 

\begin{table}[htbp]
	\resizebox{8.9cm}{!}{
		\begin{tabular}{l|c|c|c|c}
			& \begin{tabular}[c]{@{}c@{}}Autoregressive\\ (eg. PixelRNN)\end{tabular} & \begin{tabular}[c]{@{}c@{}}Variational\\ (VAEs)\end{tabular} & \begin{tabular}[c]{@{}c@{}}Adverserial\\ (GANs)\end{tabular} & \begin{tabular}[c]{@{}c@{}}NSN \\ (Ours)\end{tabular} \\ \hline
			\begin{tabular}[c]{@{}l@{}}Lower model \\ complexity?\end{tabular} & \cmark & \xmark & \xmark & \cmark \\ \hline
			\begin{tabular}[c]{@{}l@{}}Non-parametric \\ modelling?\end{tabular} & \xmark & \xmark & \xmark & \cmark \\ \hline
			Stable training? & \cmark & \cmark & \xmark & \cmark \\ \hline
			\begin{tabular}[c]{@{}l@{}}Parallel pixel\\ generation?\end{tabular} & \xmark & \cmark & \cmark & \cmark \\ \hline
			\begin{tabular}[c]{@{}l@{}}Sampling from \\ hidden layers?\end{tabular} & \xmark & \xmark & \xmark & \cmark
		\end{tabular}
	}
	\caption{ Table to compare the 
	characteristics of the proposed NSN with the 
	existing deep generative models.}
	\label{table:comparison}
\end{table}

Table \ref{table:comparison} provides a comparison of the characteristics of 
our proposed NSN with existing deep generative models. In terms of model complexity, NSN requires to maintain only a single network 
constructed with Gaussian-like filters, whereas GANs or VAEs require two or more neural networks. Furthermore, PixelRNN, VAE and GAN all are constructed 
as a parametric network. In contrast, NSNs support non-parametric training that can automatically detect 
the required number of filters at each layer of the network. The NSNs are trained with a variant of the Expectation 
Maximization (EM) algorithm  which has been shown to have a direct and stable density estimation process \cite{Bishop}.
In addition, our generation algorithm, NSN-Gen, can be applied at any 
hidden layer to visualize the image patches generated at that layer, thus enabling fine tuning of the network.

\section{Normal Similarity Network}
\label{sec:model}
The NSN architecture is inspired from CNN-based models where the layers are constructed with learnable Gaussian-style filters \cite{cnn2,cnn1}.
The inputs are image patches obtained by sliding a window over an image with some stride step.
We represent each image patch as a multi-dimensional array  $\bm{x}_i, i = 1, 2, \cdots$.
At each layer, we compute the
similarity scores of the image patches to the filters, and 
apply a sigmoid activation function to ensure that the outcomes fall in the range $[0,1]$.
These sigmoid activation maps of similarity scores are the outputs of each layer.
Note that, the output of the final layer is one-dimensional feature maps (refer to Figure \ref{fig:NSN}).

We use a variant of Expectation-Maximization (EM) algorithm at each layer to learn the filters by
maximizing the log-likelihood of the image patches \cite{Bishop}.
Our non-parametric EM is a hard clustering model 
that automatically determines the required number of filters at each layer. 
We start  with one filter  $g_1(\bm{\mu}_1,\sigma_1)$ where $\bm{\mu}_1$ and $\sigma_1$ are randomly initialized,
and iteratively create new filters as follows. Each iteration consists of an E-Step and an M-Step.

In the E-Step, we compute the similarity of 
 each image patch $\bm{x}_i$ to the filter $g_j(\bm{\mu}_j,\sigma_j)$,  $1 \leq j \leq N$, where $N$ is the number of existing filters. 
The similarity score is given as:
\begin{equation} \label{GCN-sim}
\begin{split}
Sim(\bm{x}_i,g_j(\bm{\mu}_j,\sigma_j)) = \log \Bigg( \frac{1}{\sqrt[]{2\pi}\sigma_j} \exp -\frac{||\bm{x}_i - \bm{\mu}_j||^2}{2\sigma_j^2} \Bigg) ,
\end{split} \end{equation}
where  $||\cdot||$ is the $L2$-norm. The filter mean, $\bm{\mu}_j$ is a tensor of same dimension as $\bm{x}_i$ and $\sigma_j \in \mathbb{R}$ is a scalar value.
We assign the image patch to the filter with the highest similarity score.
If this score is lower than some pre-defined threshold $\alpha$, we assign the patch to a new filter.

In the M-step, we update the parameters of  each filter $g_j$:
\begin{equation} \label{crp-M}
\bm{\mu}_j^{new} = \frac{1}{|S|}~~{\sum_{\bm{x}_i \in S} \bm{x}_i}  ~~,
\end{equation}
\begin{equation}
\sigma_j^{new} =  \sqrt {\frac{1}{|S|}
~{\sum_{\bm{x}_i \in S} ||\bm{x}_i - \bm{\mu}_j^{new}||^2} } ~~,
\end{equation}
where $S$ be the set of patches assigned to filter $g_j$. 
The algorithm terminates when converged or the number of iterations exceeds the maximum limit. 
\section{Image Generation from NSN}
\label{sec:NSNGen}
\begin{figure*}[t]
\center
\includegraphics[scale=0.4]{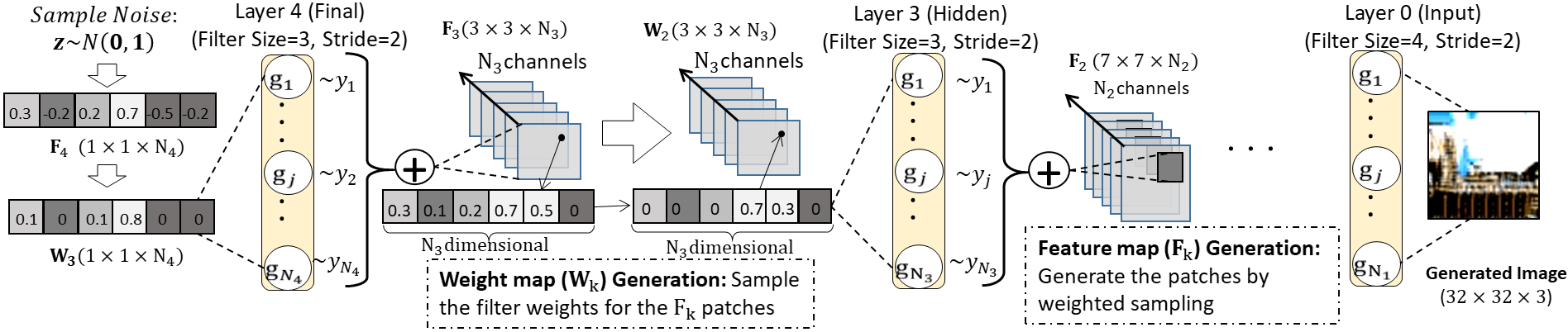}
\caption{Example of image generation process from a 4-layered NSN: 
Sample the final layer feature map, $\bm{F}_4$ from $\mathcal{N}(\bm{0},\bm{1})$. 
NSN-Gen then constructs the weight map $\bm{W}_3$, 
for the patches of $\bm{F}_3$.
We then generate the patches by combining samples drawn from $4^{th}$ layer filters.
The process continues until we reach in the input layer to generate the image.
}
\label{fig:NSN-Gen}
\end{figure*}
The final layer of NSN captures the density of downsampled training images in an one-dimensional vector space.
Our proposed image generation algorithm, NSN-Gen takes a random vector from that vector space as input and generates a new image through a backward pass.

Let us assume, there is $L$ layers in NSN and the number of filters at layer $k$ is denoted as $N_k$. Let $\bm{F}_k$ be the feature maps at layer $k$.
We start the process from layer $L$ with $\bm{F}_L$ as a random $1 \times 1 \times N_L$ dimensional noise sampled from a normal distribution $\mathcal{N}(\bm{0},\bm{1})$. 
NSN-Gen iteratively reconstructs the feature maps while progressing towards the input layer to generate an image (refer to Figure \ref{fig:NSN-Gen}).

After reconstructing $\bm{F}_k$, $\bm{F}_{k-1}$ is produced by generating its patches as a weighted combination of the samples drawn from the $N_k$ filters of layer $k$.
The weights of the samples corresponding to the $(r,c)^{th}$ patch is obtained by processing the vector $\bm{v}$, at $(r,c)^{th}$ entry of $\bm{F}_k$.
We increase the dynamic range of $\bm{v}$ by an element-wise exponential function,
$\tilde{\bm{v}} \propto \exp ({\delta_1 \bm{v}})$ followed by a normalization, where $\delta_1$ is a hyper-parameter.
We construct the weight vector $\bm{w}$ by sampling $n ~ (<< N_k)$ filter indices of layer $k$ {\em with replacement} policy and normalizing the sample counts: 
\begin{equation}\label{eqn:weightMap}
\begin{split}
\bm{w} \sim Multinomial( n, \tilde{\bm{v}}) ~~,
\end{split}
\end{equation}
where $\tilde{\bm{v}}$ represents the probability distribution of the multinomial distribution. 
We store $\bm{w}$ into the $(r,c)^{th}$ position of a
weight map $\bm{W}_{k-1}$.
Now, we draw samples $\bm{Y} = [\bm{y}_1, \dots , \bm{y}_{N_k}]$ from filters $g_j(\bm{\mu}_j, \sigma_j), 1\leq j\leq N_k$:
\begin{equation}\label{eqn:filtMap}
\begin{split}
\bm{y}_j = \bm\mu_j + \delta_2 \sigma_j \bm{M} ~~,
\end{split}
\end{equation}
where $\bm{M} \sim \mathcal{N}(\bm{0},\bm{1})$ is a random matrix of the same dimension as $\bm{\mu}_j$.
Finally, we generate the patch ($\delta_3 \textbf{w}^T\bm{Y}$) as the weighted sum of these samples, where $\delta_2$ and $\delta_3$ are hyper-parameters. 
After obtaining the patches for all $(r,c)$ positions, we stitch them together by averaging the overlapped regions to obtain $\bm{F}_{k-1}$.
Similarly, this method can be applied to draw samples from a hidden layer $k$.   
Starting with an $1\times 1 \times N_k$ dimensional noise vector 
at layer $k$ the same process is repeated till the input layer.
As NSN-Gen can be applied at any hidden layer to reconstruct the upper 
layer feature maps, it allows us to applied to a wide range of computer vision problems as demonstrated in the next section.
\section{Experiments}
\label{sec:results}
We use four different datasets in our experiments: 
MNIST \cite{mnist} and Fashion \cite{mnist-fashion} 
contains $28\times28$ gray scale images of 
hand written digits and cloths. 
LFW \cite{LFW} and Church \cite{lsun} are RGB images of human faces 
and church buildings that are resized to $64\times64$.
For the LFW dataset, we preprocess the images with channel-wise ZCA transformation as this enables our network to generate sharper human face images.
 We train 3-layer NSNs for MNIST and Fashion 
 and 5-layer NSNs for LFW and Church datasets.
 The required number of filters at each layer are automatically determined 
 during training. 
\subsection{Comparative Study}
\begin{figure}[ht]
\centering
\begin{subfigure}{.2\textwidth}
  \centering
  \includegraphics[scale=0.5]{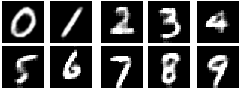}
  \caption{MNIST Samples}
  \label{fig:gen1}
\end{subfigure}%
\begin{subfigure}{.2\textwidth}
  \centering
  \includegraphics[scale=0.5]{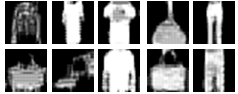}
  \caption{Fashion Samples}
  \label{fig:gen2}
\end{subfigure}
\caption{ Visualization of generated samples from NSN.}
\label{fig:gen28}
\end{figure}

\begin{figure}[htbp]
\centering
\begin{subfigure}{.5\textwidth}
  \centering
  \includegraphics[scale=0.47]{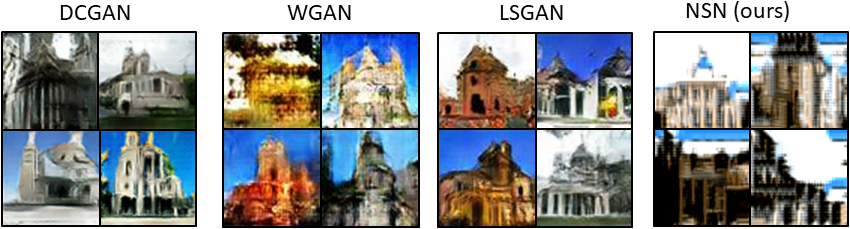}
  \caption{Generated Church samples }
  \label{fig:lsgan64}
\end{subfigure}%

\begin{subfigure}{.5\textwidth}
  \centering
  \includegraphics[scale=0.47]{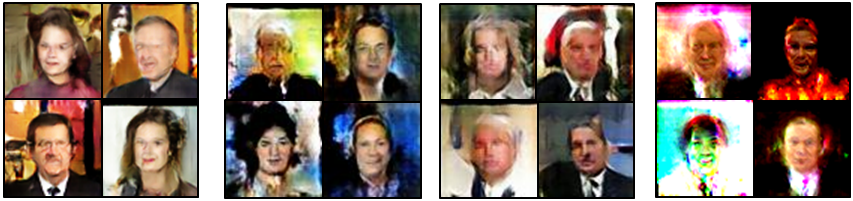}
  \caption{Generated LFW samples }
  \label{fig:nsn64}
\end{subfigure}
\caption{ Visualization of the best samples generated by different 
state-of-the-art GAN models and our proposed NSN.}
\label{fig:gen64}
\end{figure}

 We carry out a qualitative comparison of the  samples generated by the proposed NSNs with state-of-the-art GAN models.
 We compute the inception score with 50,000 samples randomly generated by these models \cite{improvedGan}.
A higher Inception Score is achieved if a pre-trained
inception convolutional network classifies the samples with higher
certainty as well as identifies a diverse variety among them.
Table \ref{table:inception} shows that  the samples generated 
from NSN are comparable with the state-of-the-art models.
Figure \ref{fig:gen28} shows the NSN-generated samples of 
MNIST and Fashion.
Figure \ref{fig:gen64} presents the Church and LFW samples 
generated by DCGAN \cite{dcgan2016}, WGAN \cite{wgan}, 
LSGAN \cite{lsgan} and our NSN for visual comparison.
\begin{table}[htbp]
	\centering
	\resizebox{7.5cm}{!}{
		\begin{tabular}{|l|l|l|l|l|}
			\hline
			& MNIST & Fashion & LFW & Church \\ \hline
			{\begin{tabular}[c]{@{}l@{}}Real Data\\ \end{tabular} }&
			1.99 $\pm$0.01 & 4.25$\pm$0.05 & 3.64$\pm$0.28 & 2.70$\pm$0.04 \\ \hline
			
			{DCGAN \cite{dcgan2016}} &  
			2.27$\pm$0.01 & 4.15$\pm$0.04 & 2.67$\pm$0.03 & 2.89 $\pm$ 0.02 \\ \hline
			
			{WGAN \cite{wgan}} &
			2.02$\pm$0.02 & 3.00$\pm$0.03 & 2.94$\pm$0.03 & 3.05$\pm$ 0.02 \\ \hline
			
			{LSGAN \cite{lsgan}} &
			2.03$\pm$0.01 & \textbf{4.45$\pm$0.03} & 2.33$\pm$0.02 & \textbf{3.44$\pm$ 0.04} \\ \hline
			
			\textbf{NSN (Ours)} & 
			\textbf{2.44$\pm$0.02} & 3.85$\pm$0.04 & \textbf{3.04$\pm$0.10} & 2.99$\pm$ 0.02 \\ \hline
		\end{tabular}
	}
	\caption{Inception score calculated on 50,000 images randomly sampled from GAN models and our proposed NSNs. Score for LFW(Real Data) is calculated for entire dataset of 13,233 images. Higher scores are better.}
	\label{table:inception}
\end{table}
\subsection{Effect of NSN-Gen hyper-parameters}
\begin{figure}[htbp]
 \centering
 \begin{subfigure}{.5\textwidth}
   \centering
   \includegraphics[scale=0.52]{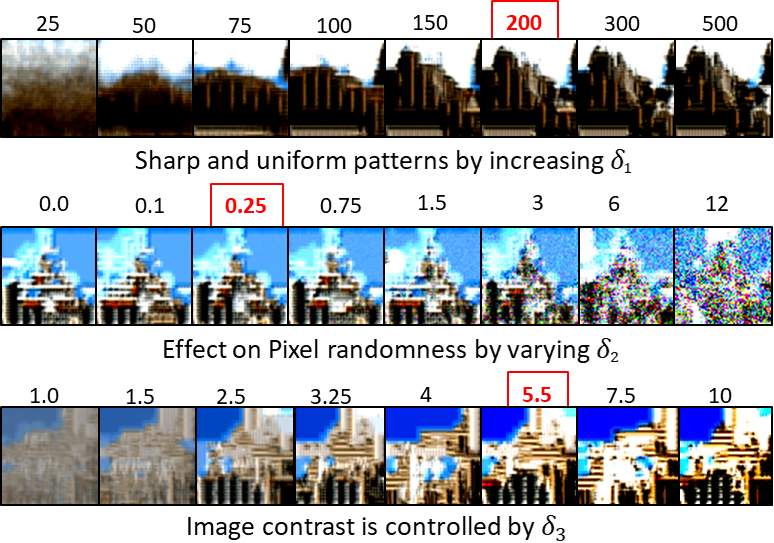}
   \caption{Generated Church samples}
   \label{fig:churchOmega}
 \end{subfigure}
 \begin{subfigure}{.5\textwidth}
   \centering
   \includegraphics[scale=0.52]{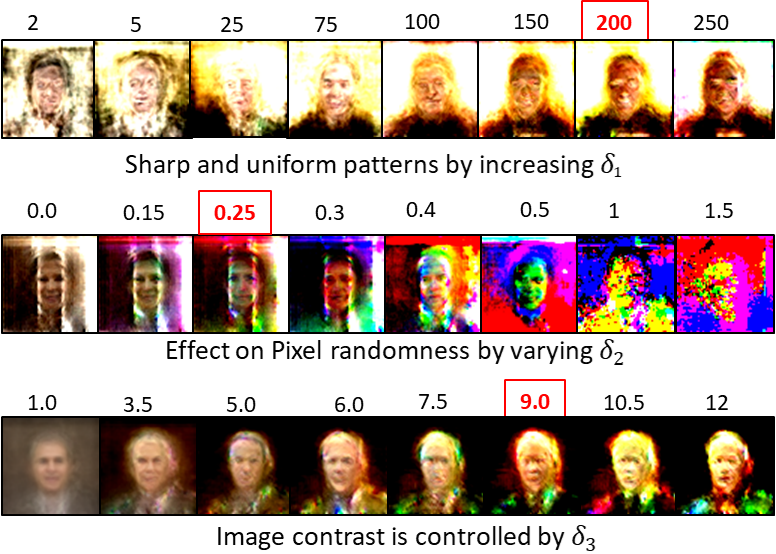}
   \caption{Generated LFW Samples}
   \label{fig:ifwOmega}
 \end{subfigure}%
 \caption{ Visualizing the effect of NSN-Gen hyper-parameters ($\delta_1, \delta_2, \delta_3$) on the generated images. Values in the red boxes presents
 the selected hyper-parameters for computing the inception scores.
 We vary one of the three hyper-parameters at each row to produce the 
 images from the same noise vector.
 }
 \label{fig:omega}
\end{figure}

We use the same values for the hyper-parameters ($\delta_1, \delta_2$ and $\delta_3$) across different layers for NSN-Gen. 
They are selected to maximize the inception score.
We can control the characteristics of generated images by 
these hyper-parameters.
Figure \ref{fig:omega}, shows the generated images as we vary one of them while keeping same values for the others.
As we increase $\delta_1$, it increases the dynamic range of filter similarity scores.
As a result, choosing the filter with maximum similarity score becomes more probable (Eq \ref{eqn:weightMap}), resulting generation of uniform and sharper images. However, a very large value for $\delta_1$ reduces the diversity among the generated patterns and thereby reduces the inception score.
$\delta_2$ influences the pixel randomness while $\delta_3$ controls the contrast for the generated images.

\subsection{Applications of NSNs}

In this section, we show how NSN can be also used for image styling and reconstruction of occluded images.
For these applications, we train NSNs with images preprocessed with channel-wise ZCA whitening. 
This preprocessing is useful to remove correlations which may exist
among the neighborhood pixels in natural images.
This allows us to obtain evenly distributed filters over the input patches \cite{learningKmeans}.

\begin{figure}[htbp]
 	\centering
 	\begin{subfigure}{.22\textwidth}
   	\centering
   	\includegraphics[scale=0.45]{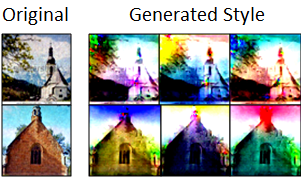}
   \caption{Church images}
   \label{fig:style2church}
 \end{subfigure}
 \begin{subfigure}{.22\textwidth}
   \centering
   \includegraphics[scale=0.45]{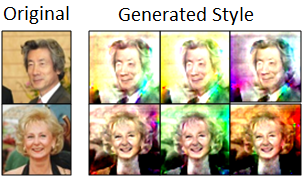}
   \caption{LFW images}
   \label{fig:style2lfw}
 \end{subfigure}%
 	\caption{Example of generated style images using NSNs.}
 	\label{fig:style}
 \end{figure}

\smallskip
\noindent\textbf{Image Styling.} 
Image styling is an attractive and emerging field in computer vision 
with many successful industrial applications \cite{neuralStyle}.
As we downsample an image into the NSN and reconstruct back 
from its first layer feature map, 
the network adds a fractal style in the image.  
 Figure \ref{fig:style} shows that NSN-Gen is able to generate many different style images for the same input image
  due to its inherent sampling properties.

\begin{figure}[htbp]
 	\centering
 	\begin{subfigure}{.24\textwidth}
   	\centering
   	\includegraphics[scale=0.43]{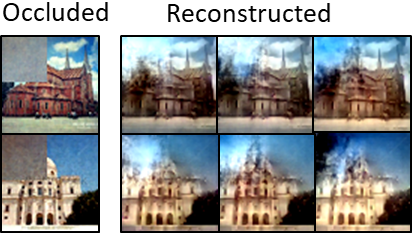}
   \caption{Church images}
   \label{fig:inpainting2church}
 \end{subfigure}
 \begin{subfigure}{.22\textwidth}
   \centering
   \includegraphics[scale=0.43]{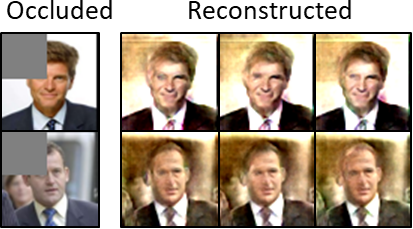}
   \caption{LFW images}
   \label{fig:inpainting2lfw}
 \end{subfigure}%
 	\caption{Example of occluded image reconstruction by NSN.}
 	\label{fig:inpainting}
 \end{figure}

\smallskip
\noindent\textbf{Reconstruction of Occluded Images.}
As the final layer of an NSN captures the density in a transformed training domain, the final layer feature maps of an occluded image should 
represent feature maps of a valid image.
Here, we aim to reconstruct the occluded portion by conditioning on this final layer feature map.
We start with this final layer feature map and apply NSN-Gen to reconstruct the occluded portion in its first layer feature map.
The image is then reconstructed from this modified first-layer feature map that appropriately fills the occluded part as shown in Figure \ref{fig:inpainting}. 
Note that, in many cases, this procedure looses the color information due to the ZCA preprocessing and reconstruction errors introduced by NSN-Gen.
However, we can add colors and style on these generated images by adjusting the NSN-Gen hyper-parameters.

\section{Conclusion}
\label{sec:conclusion}

Generative modelling with NSN presents an encouraging direction for unsupervised image modelling that aims to learn the density function of the final layer activation maps of the training images by maximizing the likelihood. The proposed NSN-Gen presents a technique to generate images from a noise vector by reconstructing activation maps at each layers. Experiments suggest that once the model is trained, it can be adopted for various computer vision applications.

{
\bibliographystyle{IEEEbib}
\bibliography{IEEEbib}
}

\newpage
\section{Additional Results}

\subsection{NSN Architecture for Image Generation}

Table \ref{table:param28} and Table \ref{table:param64} respectively presents the 
NSN architectures used for $(28 \times 28)$ MNIST \cite{mnist} and Fashion 
\cite{mnist-fashion} and $(64 \times 64)$ LFW \cite{LFW} and Church \cite{lsun} datasets.
At each layer, we apply stride step$=2$ to downsample the feature maps to (approximately) half of its size.
Note that, our non-parametric EM algorithm automatically determines the required number of filters during training at each layer of the network.

\begin{table}[ht]
\centering
\small
\begin{tabular}{|l|l|l|l|l|}
\hline
\multirow{2}{*}{Layers} & \multirow{2}{*}{Input} & \multirow{2}{*}{Filter, Stride} & \multicolumn{2}{l|}{\begin{tabular}[c]{@{}l@{}}No. of filters automatically \\ determined during training \end{tabular}} \\ \cline{4-5} 
 &  &  & MNIST & \begin{tabular}[c]{@{}l@{}}MNIST:\\ Fashion\end{tabular} \\ \hline
L0 & 28$\times$ 28 & (4$\times$4),~2 & 96 & 140 \\ \hline
L1 & 13$\times$ 13 & (3$\times$3),~2 & 130 & 238 \\ \hline
L2 & 6 $\times$ 6  & (6$\times$6),~2 & 131 & 256 \\ \hline
\end{tabular}
\caption{Description of 3-layered NSN architecture applied for $(28 \times 28)$ images of MNIST and Fashion datasets. }
\label{table:param28}
\end{table}

\begin{table}[ht]
\centering
\small
\begin{tabular}{|l|l|l|l|l|}
\hline
\multirow{2}{*}{Layers} & \multirow{2}{*}{Input} & \multirow{2}{*}{Filter, Stride} & \multicolumn{2}{l|}{\begin{tabular}[c]{@{}l@{}}No. of filters automatically \\ determined during training \end{tabular}} \\ \cline{4-5} 
 &  &  & LFW & \begin{tabular}[c]{@{}l@{}}LSUN:\\ Church\end{tabular} \\ \hline
L0 & 64 $\times$ 64 & (4$\times$4),~2 & 284 & 273 \\ \hline
L1 & 31 $\times$ 31 & (3$\times$3),~2 & 242 & 263 \\ \hline
L2 & 15 $\times$ 15 & (3$\times$3),~2 & 240 & 325 \\ \hline
L3 & 7 $\times$ 7 & (3$\times$3),~2 & 235 & 398 \\ \hline
L4 & 3 $\times$ 3 & (3$\times$3),~1 & 161 & 210 \\ \hline
\end{tabular}
\caption{Description of 5-layered NSN architecture applied for 
$(64 \times 64)$ images of LFW and Church datasets.} 
\label{table:param64}
\end{table}

\begin{figure}[ht]
\center
\includegraphics[scale=0.54]{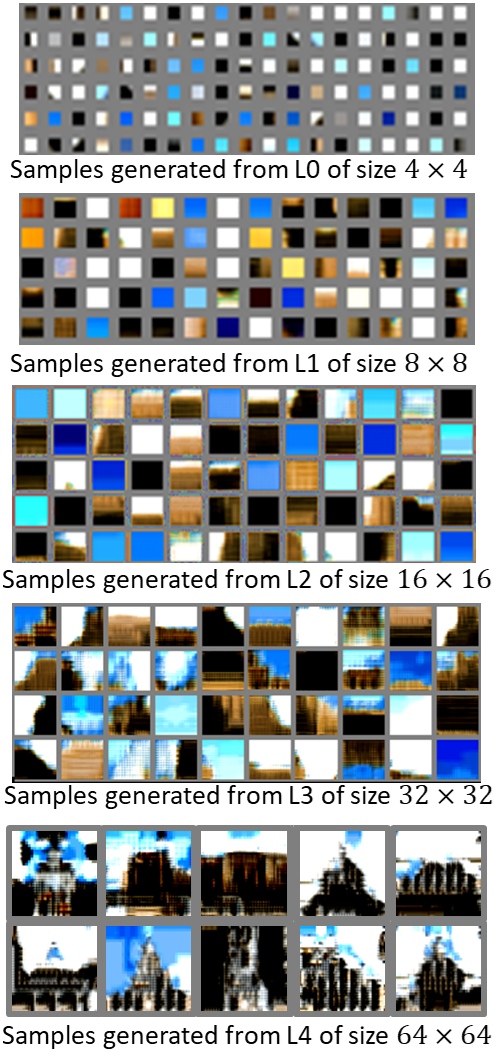}
\caption{Visualization of samples generated from the hidden layers of 
NSN trained on the Church \cite{lsun} images. 
}
\label{fig:layerSamples}
\end{figure}

\subsection{Sampling from Intermediate layers}
NSN-Gen can be applied at any hidden layer to visualize samples at different hidden layers of the network.
This can be useful to fine tune the network parameters as well as
the convergence of the training algorithms.
To generate image patches from layer $k$, we sample a 
$1\times 1 \times N_k$ dimensional 
noise vector from $\mathcal{N}(\textbf{0},\textbf{1})$;
where $N_k$ represents the number of filters at layer $k$.
This noise vector is represented as a patch for the feature map $\bm{F}_k$ of layer $k$ .
Now, we start from layer $k$ and apply NSN-Gen to generate an image patch in the input layer.
Figure \ref{fig:layerSamples} shows the generated samples from different 
intermediate layers of NSN, trained on Church dataset \cite{lsun}.

\begin{figure*}[htb]
\centering
\begin{subfigure}{.5\textwidth}
  \centering
  \includegraphics[scale=0.65]{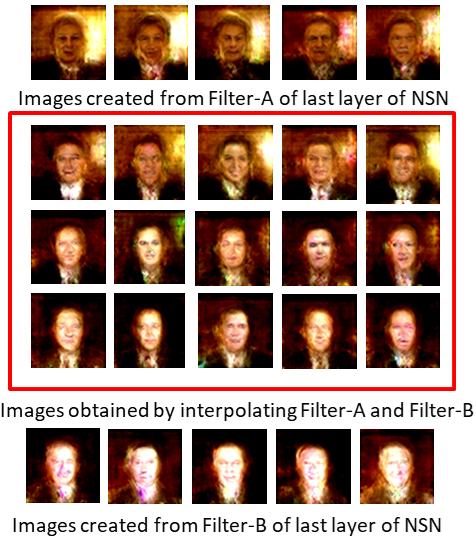}
  \caption{Interpolation between two noise vectors corresponding to two different filters.}
  \label{fig:interpolate}
\end{subfigure}%
\begin{subfigure}{.5\textwidth}
  \centering
  \includegraphics[scale=0.55]{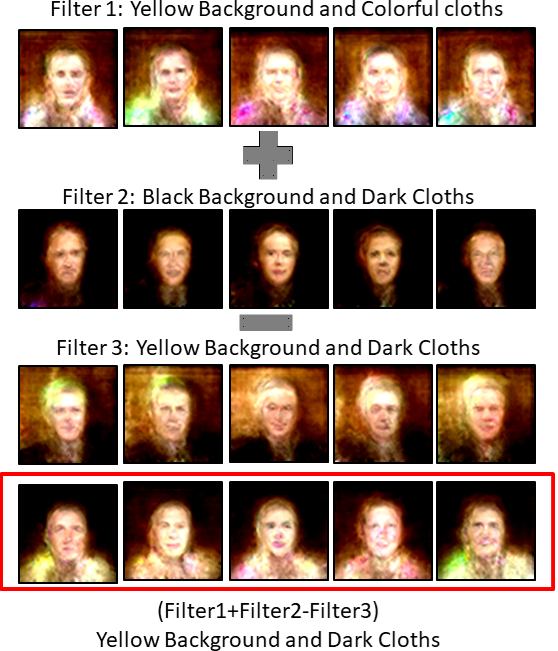}
  \caption{Arithmetic operation among the final layer NSN filters.}
  \label{fig:arithmetic}
\end{subfigure}

\caption{
(a) Top and bottom rows are showing the images generated from two different filters.
While Filter-A (top row) is generating darker faces in a yellowish background, 
Filter-B (bottom row)is generating brighter faces in black background.
As we can see in the red box, how the attributes of the images are changing 
as we move from Filter-A to Filter-B by interpolating the noise vector.
~\\
(b) Top three rows are showing the generated images from three different filters. 
To add the features of two filters and subtract it from the third filter, 
we generate the final-layer input feature map (i.e $\bm{F}_3$) 
and apply the desired arithmetic operation. 
The resultant feature map is then used to generate the image using NSN-Gen. 
As we can see, the images generated in the last row are keeping the features 
of Filter 1 and Filter 2 while throwing away the features of Filter 3.
}
\label{fig:zOperation}
\end{figure*}

\subsection{Manipulating Noise Vectors for Image Generation}
Similar to GANs, we can also manipulate the noise vectors $z$,
to manipulate generated images from NSN, revealing rich linear structure of visual concepts established in the noise space \cite{dcgan2016}.
Further, we can construct a noise vector to generate image 
from a particular final layer filter of NSN.
To generate an image from a filter $g_j$, we encode the noise as a one-hot vector with all except the $j^{th}$ index as zero and apply NSN-Gen.
Note that, images generated from different filters have different visual characteristics.
As we can see in Figure \ref{fig:interpolate}, by interpolating the noise vectors corresponding to two different filters, we can generate new images with interpolated visual properties of them.
Note that, when NSN is trained with ZCA per-processed images, due to the inherited sampling properties of NSN-Gen, we obtain different images 
with similar visual properties as we repeatedly generate samples from the same noise vector. 

Further, we can apply simple arithmetic on the final layer NSN filters to generate new images of combined properties (Figure \ref{fig:arithmetic}). 
After choosing the noise vectors corresponding to the filters, we apply 
NSN-Gen to generate the final-layer input feature maps 
(i.e $\bm{F}_{L-1}$) corresponding from the noise vectors.
Now, we apply our desired arithmetic operation on these feature maps and reconstruct the images from the new feature maps. 
 
\begin{figure*}[htbp]
\centering
\begin{subfigure}{.5\textwidth}
  \centering
  \includegraphics[scale=0.7]{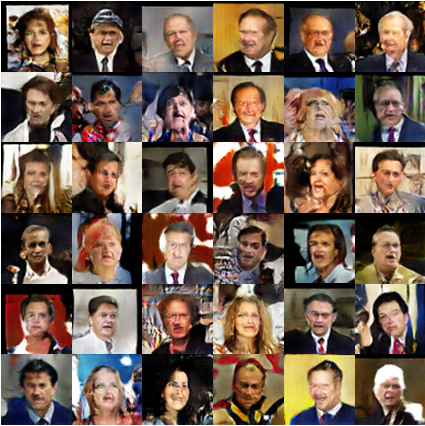}
  \caption{DCGAN Samples}
  \label{fig:dcgan64h}
\end{subfigure}%
\begin{subfigure}{.5\textwidth}
  \centering
  \includegraphics[scale=0.7]{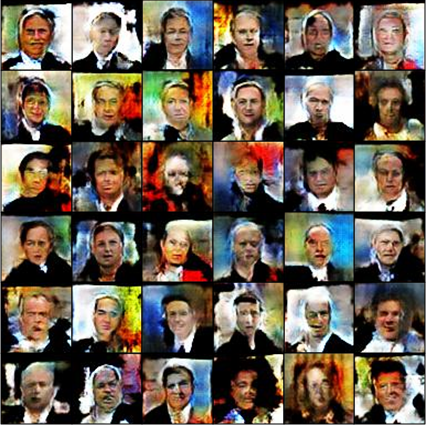}
  \caption{WGAN Samples}
  \label{fig:wgan64h}
\end{subfigure}

\begin{subfigure}{.5\textwidth}
  \centering
  \includegraphics[scale=0.7]{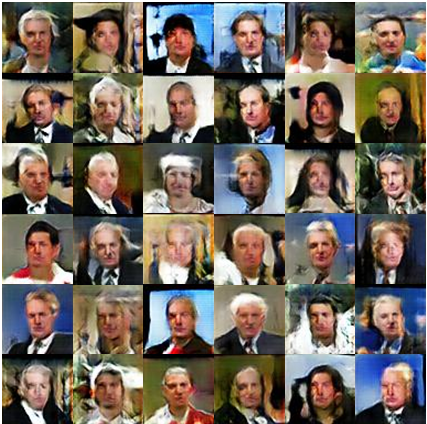}
  \caption{LSGAN Samples}
  \label{fig:lsgan64h}
\end{subfigure}%
\begin{subfigure}{.5\textwidth}
  \centering
  \includegraphics[scale=0.7]{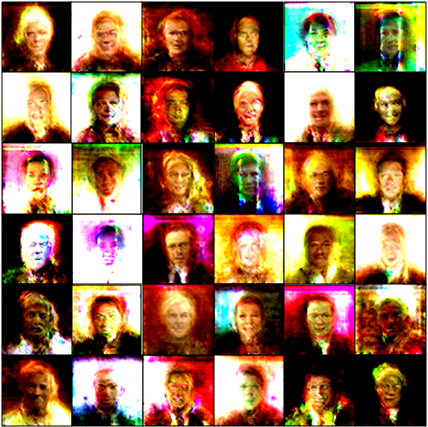}
  \caption{NSN(ours) Samples}
  \label{fig:nsn64h}
\end{subfigure}
\caption{ More examples of generated samples by different 
state-of-the-art GAN models and our proposed NSN trained on 
LFW dataset \cite{LFW}. }
\label{fig:gen64h}
\end{figure*}

\begin{figure*}[htbp]
\centering
\begin{subfigure}{.5\textwidth}
  \centering
  \includegraphics[scale=0.7]{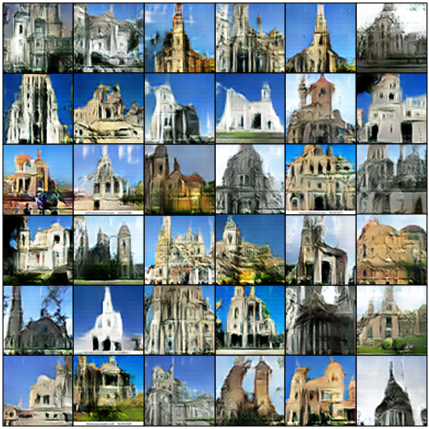}
  \caption{DCGAN Samples}
  \label{fig:dcgan64c}
\end{subfigure}%
\begin{subfigure}{.5\textwidth}
  \centering
  \includegraphics[scale=0.7]{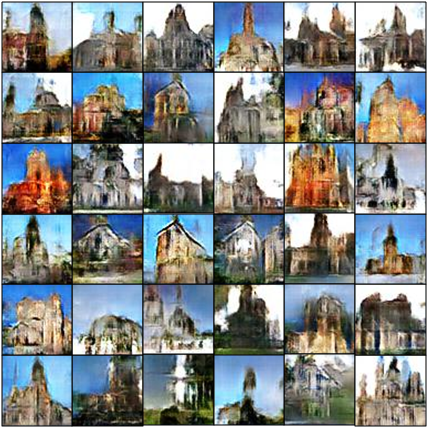}
  \caption{WGAN Samples}
  \label{fig:wgan64c}
\end{subfigure}

\begin{subfigure}{.5\textwidth}
  \centering
  \includegraphics[scale=0.7]{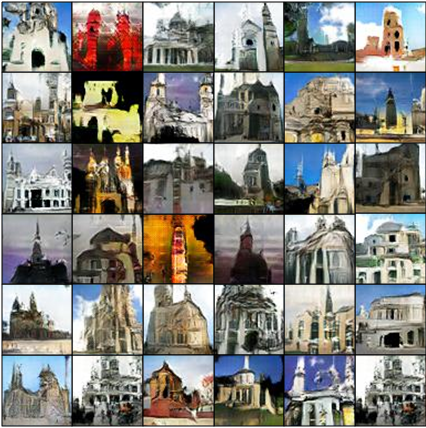}
  \caption{LSGAN Samples}
  \label{fig:lsgan64c}
\end{subfigure}%
\begin{subfigure}{.5\textwidth}
  \centering
  \includegraphics[scale=0.7]{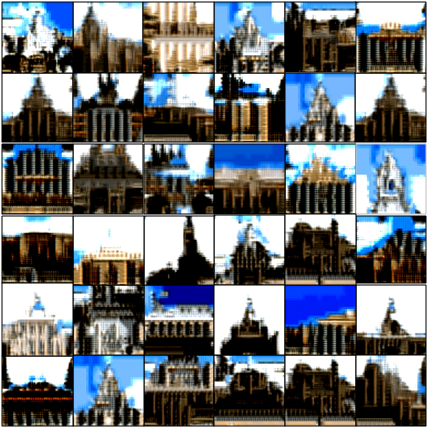}
  \caption{NSN(ours) Samples}
  \label{fig:nsn64c}
\end{subfigure}
\caption{ More examples of generated samples by different 
state-of-the-art GAN models and our proposed NSN trained on 
Church dataset \cite{lsun}.  }
\label{fig:gen64c}
\end{figure*}

\begin{figure*}[t]
\centering
\begin{subfigure}{.5\textwidth}
  \centering
  \includegraphics[scale=0.63]{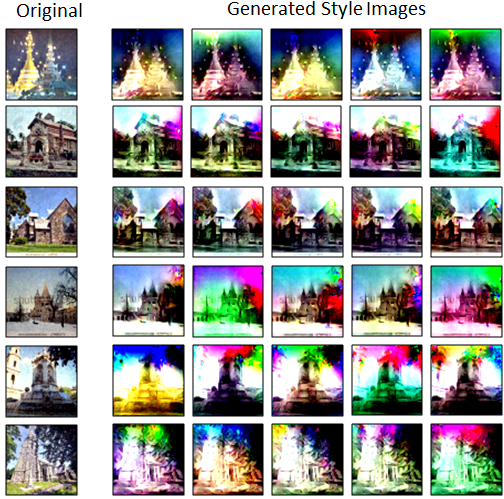}
\end{subfigure}%
\begin{subfigure}{.5\textwidth}
  \centering
  \includegraphics[scale=0.63]{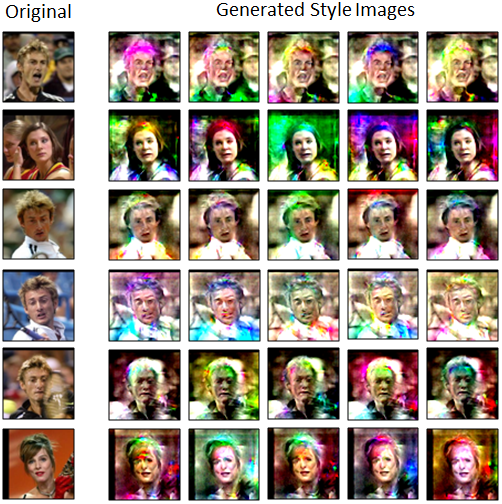}
\end{subfigure}

\caption{ More examples of generated style images from the proposed NSN. 
Left: Church \cite{lsun}, Right: LFW \citep{LFW}.
}
\label{fig:style64}
\end{figure*}

\begin{figure*}[hbpt]
\center
\includegraphics[scale=0.66]{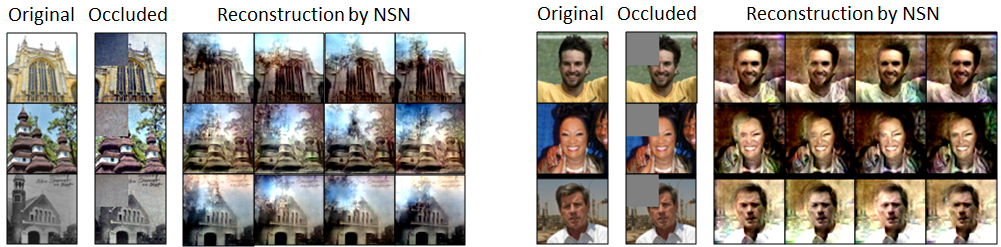}
\caption{ More example of image reconstruction from occluded images by NSN.}
\label{fig:inpaint4}
\end{figure*}

\end{document}